\documentclass[11pt,a4paper]{article}
\usepackage[hyperref]{acl2021}
\usepackage{times}
\usepackage{latexsym}
\usepackage{xcolor}
\usepackage{tcolorbox}
\usepackage{amsmath}
\usepackage{amssymb}
\usepackage{bbm}
\usepackage{bm}

\usepackage{microtype}

\tcbset{
    frame code={}
    center title,
    left=0pt,
    right=0pt,
    top=0pt,
    bottom=0pt,
    colback=gray!40,
    colframe=white,
    enlarge left by=0mm,
    enlarge bottom by=-5mm,
    boxsep=3pt,
    arc=0pt,outer arc=0pt,
    }

\definecolor{textgray}{gray}{0.8}
\aclfinalcopy 

\title{Ultra-Fine Entity Typing with Weak Supervision from a Masked Language Model}

\author{Hongliang Dai, Yangqiu Song \\
  Department of CSE, HKUST \\
  \texttt{\{hdai,yqsong\}@cse.ust.hk} \\\And
  Haixun Wang \\
  Instacart \\
  \texttt{haixun.wang@instacart.com} \\}

\date{}

\begin{document}
\maketitle
\begin{abstract}

Recently, there is an effort to extend fine-grained entity typing by using a richer and ultra-fine set of types, and labeling noun phrases including pronouns and nominal nouns instead of just named entity mentions. A key challenge for this ultra-fine entity typing task is that human annotated data are extremely scarce, and the annotation ability of existing distant or weak supervision approaches is very limited.
To remedy this problem, in this paper, we propose to obtain training data for ultra-fine entity typing by using a BERT Masked Language Model (MLM). Given a mention in a sentence, our approach constructs an input for the BERT MLM so that it predicts context dependent hypernyms of the mention, which can be used as type labels. Experimental results demonstrate that, with the help of these automatically generated labels, the performance of an ultra-fine entity typing model can be improved substantially. We also show that our approach can be applied to improve traditional fine-grained entity typing after performing simple type mapping.
\end{abstract}

\section{Introduction}
Fine-grained entity typing \cite{ling2012fine} has been long studied in the natural language processing community as the extracted type information is useful for downstream tasks such as entity linking \cite{ling2015design,onoe2020interpretable}, relation extraction \cite{koch2014type}, coreference resolution \cite{onoe2020interpretable}, etc.
Recently, ultra-fine entity typing~\cite{choi2018ultra} extends the effort to using a richer set of types (e.g., \textit{person}, \textit{actor}, \textit{company}, \textit{victim}) to label noun phrases including not only named entity mentions, but also pronouns and nominal nouns.
This task directly uses type words or phrases as tags. Its tag set can contain more than 10,000 types.
A challenge is that with the large type set, it is extremely difficult and time-consuming for humans to annotate samples. As a result, most existing {works use weak labels that are automatically generated} \cite{ling2012fine,choi2018ultra,lee2020chinese}. 

There are two main approaches to obtaining weakly labeled training {examples}. One approach is to find the Wikipedia pages that correspond to entity mentions, which can be done by using hyperlinks to Wikipedia or applying entity linking. Then the entity types can be obtained from knowledge bases. The other approach is to directly use the head words of nominal mentions as ultra-fine type labels.
For example, if a nominal mention is ``a famous actor,'' then the head word ``actor'' can be used as its type label.

\begin{table*}
\centering
\begin{tabular}{p{10.2cm}p{4.7cm}}
\hline \textbf{Input} & \textbf{Top Words for [MASK]} \\ \hline
In late 2015, \colorbox{blue!30}{[MASK] such as} \underline{\textbf{Leonardo DiCaprio}} starred in The Revenant. & actors, stars, actor, directors, filmmakers \\
\hline
At some clinics, \underline{\textbf{they}} \colorbox{blue!30}{and some other [MASK]} are told the doctors don't know how to deal with AIDS, and to go someplace else. & patients, people, doctors, kids, children \\ \hline
Finkelstein says he expects the company to ``benefit from some of the disruption faced by \underline{\textbf{our competitors}} \colorbox{blue!30}{and any other [MASK]}.'' & company, business, companies, group, investors \\ \hline
\end{tabular}
\caption{\label{tab:weak-examples} Examples of constructed BERT MLM inputs for obtaining weak entity typing labels. Entity mentions are in bold and underlined. The texts highlighted with blue background are not in the original sentences. They are inserted to create inputs for BERT. The right column lists the five most probable words predicted by a pretrained BERT-Base-Cased MLM.}
\end{table*}

{Several problems exist when using these weak labels for the ultra-fine typing task.}
First, in the dataset created by \citet{choi2018ultra}, on average there are fewer than two labels (types) for each sample annotated through either entity linking or head word supervision. On the other hand, a human annotated sample has on average 5.4 labels. As a result, models trained from the automatically obtained labels have a low recall. Second, neither of the above approaches can create a large number of training samples for pronoun mentions. Third, it is difficult to obtain types that are highly dependent on the context. For example, in ``I met the movie star Leonardo DiCaprio on the plane to L.A.,'' the type \textit{passenger} is correct for ``Leonardo DiCaprio.'' However, this type cannot be obtained by linking to knowledge bases.

In this paper, to alleviate the problems above, we propose an approach that combines hypernym extraction patterns \cite{hearst1992automatic,seitner2016large} with a masked language model {(MLM), such as BERT \cite{devlin2019bert},} to generate weak labels for ultra-fine entity typing.
Given a sentence that contains a mention, our approach adds a short piece of text that contains a ``[MASK]'' token into it to construct an input {to BERT}. Then, the pretrained MLM will predict the hypernyms of the mention as the most probable words for ``[MASK].'' These words can then be used as type labels.
{For example}, consider the first {example} in Table \ref{tab:weak-examples}. The original sentence is ``In late 2015, Leonardo DiCaprio starred in The Revenant.'' We construct an input for the BERT MLM by inserting ``[MASK] such as'' before the mention ``Leonardo DiCaprio.'' With this input, the pretrained BERT MLM predicts ``actors,'' ``stars,'' ``actor,'' ``directors,'' {and} ``filmmakers'' as the five most probable words for ``[MASK].'' Most of them are correct types for the mention after singularization.
{This approach can generate labels for different kinds of mentions, including named entity mentions, pronoun mentions, and nominal mentions. Another advantage is that it can produce labels that needs to be inferred from the context. This allows us to generate more context-dependent labels for each mention, such as \textit{passenger}, \textit{patient}, etc.}







Then, we propose a method to select from the results obtained through different hypernym extraction patterns to improve the quality of the weak labels. We also use a weighted loss function to make better use of the generated labels for model training. Finally, we adopt a self-training step to further improve the performance of the model. We evaluate our approach with the dataset created by \citet{choi2018ultra}, which to the best of our knowledge, is the only English ultra-fine entity typing dataset currently available. On this dataset, we achieve more than 4\% absolute F1 improvement over the previously reported best result.
Additionally, we also apply our approach to a traditional fine-grained entity typing dataset: Ontonotes \cite{gillick2014context}, where it also yields better performance than the state of the art.

Our contributions are summarized as follows.

\begin{itemize}
\item We propose a new way to generate weak labels {for ultra-fine entity typing}.

\item We propose an approach to make use of the newly obtained weak labels to improve entity typing results.

\item We conduct experiments on both an ultra-fine entity typing dataset and a traditional fine-grained entity typing dataset to verify the effectiveness of our method.

\end{itemize}

Our code is available at \url{https://github.com/HKUST-KnowComp/MLMET}.

\section{Related Work}

The ultra-fine entity typing task proposed by \citet{choi2018ultra} uses a large, open type vocabulary to achieve better type coverage than the traditional fine-grained entity typing task \cite{ling2012fine} that uses manually designed entity type ontologies.
There are only limited studies on this newly proposed task: A neural model introduced by \cite{onoe2019learning} filters samples that are too noisy to be used and relabels the remaining samples to get cleaner labels. A graph propagation layer is introduced by \cite{xiong2019imposing} to impose a label-relational bias on entity typing models, so as to implicitly capture type dependencies. \citet{onoe2021modeling} use box embeddings to capture latent type hierarchies. There is also some work on the applications of ultra-fine entity typing: \citet{onoe2020interpretable} apply ultra-fine entity typing to learn entity representations for two downstream tasks: coreference arc prediction and named entity disambiguation.

The traditional fine-grained entity typing task \cite{ling2012fine,yosef2012hyena} is closely related to ultra-fine entity typing. Automatic annotation \cite{ling2012fine,gillick2014context,dai2020exploiting} is also commonly used in the studies of this task to produce large size training data. 
Many different approaches have been proposed to improve fine-grained entity typing performance. For example, denoising the automatically generated labels \cite{ren2016label}, taking advantage of the entity type hierarchies or type inter-dependencies \cite{chen2020hierarchical,murty2018hierarchical,lin2019attentive}, exploiting external resources such as the information of entities provided in knowledge bases \cite{jin2019fine,dai2019improving,xin2018improving}, etc.

Our work is also related to recent studies \cite{petroni2019language,jiang2020can,zhang2020empower} that probe pretrained language models to obtain knowledge or results for target tasks. Different from them, we use the predictions produced by BERT as intermediate results that are regarded as weak supervision to train better models. \cite{zhang2020empower} also uses Hearst patterns to probe masked language models. However, they target at the entity set expansion task.

\section{Methodology}
\label{sec:method}
Our methodology consists of two main steps. First, we obtain weak ultra-fine entity typing labels from a BERT masked language model. Second, we use the generated labels in model training to learn better ultra-fine entity typing models. 

\subsection{Labels from BERT MLM}
\label{sec:bert-mlm-labels}


Given a sentence and a mention of interest in the sentence, our goal is to derive the hypernym or the type of the mention using a BERT MLM. To do this, we insert into the sentence a few tokens to create an artificial Hearst pattern~\cite{hearst1992automatic}. One of the inserted tokens is a special ``[MASK]'' token, which serves as the placeholder of the hypernym of the mention. As the BERT MLM predicts the ``[MASK]'' token, we derive the hypernyms of the mention.

Consider the first sentence in Table \ref{tab:weak-examples} as an example: ``In late 2015, Leonardo DiCaprio starred in The Revenant.'' To find the hypernym or the type of ``Leonardo DiCaprio'', we insert three tokens to create a ``such as'' pattern: ``In late 2015, [MASK] such as Leonardo DiCaprio starred in The Revenant.'' Applying the BERT MLM on the sentence, we derive hypernyms such as ``actors,'' ``stars,'' ``directors,'' ``filmmakers.'' Table \ref{tab:weak-examples} shows a few more examples.

\begin{table}
\centering
\begin{tabular}{lc}
\hline \textbf{Pattern} & \textbf{F1} \\ \hline
$M$ and any other $H$ & 25.3 \\
$M$ and some other $H$ & 24.8 \\
$H$ such as $M$ & 20.7 \\
such $H$ as $M$ & 18.1 \\
$H$ including $M$ & 17.4 \\
$H$ especially $M$ & 11.5 \\
\hline
\end{tabular}
\caption{\label{tab:patterns} Hypernym extraction patterns. $M$ denotes the hyponym; $H$ denotes the hypernym. The F1 score is evaluated with the development set of the ultra-fine dataset \cite{choi2018ultra} for the labels generated with the corresponding pattern.}
\end{table}

We consider the 63 Hearst-like patterns \cite{hearst1992automatic} presented in \cite{seitner2016large} that express a hypernym-hypnonym relationship between two terms. Table \ref{tab:patterns} lists some of the patterns, wherein $H$ and $M$ denote a hypernym and a hyponym, respectively. For example, ``$M$ and some other $H$'' can be used to match ``Microsoft and some other companies.''

The general procedure to use these patterns to create input samples for BERT MLM and obtain labels from its predictions is as follows. We first regard the mention as $M$. Then, we insert the rest of the pattern either before or after the mention, and  we replace $H$ with the special ``[MASK]'' token. After applying the BERT MLM on sentences with artificial Hearst patterns, we derive top $k$ type labels from the prediction for ``[MASK].''
To drive these $k$ labels, we first sigularize the most probable words that are plural. Then, remove those that are not in the type vocabulary of the dataset. Finally, use the most probable $k$ different words as $k$ labels. For example, if we want to obtain 3 labels, and the most probable words are ``people,'' ``actors,'' ``celebrities,'' ``famous,'' ``actor,'' etc. Then the 3 labels should be \textit{person}, \textit{actor}, \textit{celebrity}. Because ``actor'' is the singluar form of ``actors,'' and ``famous'' is not in the type vocabulary.

We show the performance of our method for obtaining 10 type labels for each mention with different patterns in Table \ref{tab:patterns}. A pre-trained BERT-Base-Cased MLM is used to obtain the results\footnote{We use the pretrained model provided in the Transformers library. We also tried using BERT-Large and RoBERTa models. However, they do not yield better performance.}.


For nominal mentions, directly applying the patterns that starts with ``$M$'' with the above procedure may sometimes be problematic. For example, consider the noun phrase ``the factory in Thailand'' as a mention. If we use the ``$M$ and some other $H$'' pattern and insert ``and other [MASK]'' after the mention, the BERT MLM will predict the type \textit{country} for Thailand instead of for the entire mention. To avoid such errors, while applying patterns that starts with ``$M$'' for nominal mentions, we regard the head word of the mention as $M$ instead.


A more subtle and challenging problem is that the quality of the type labels derived from different patterns for different mentions can be very different. For example, for the mention ``He'' in sentence ``He has won some very tough elections and he's governor of the largest state,'' the pattern ``$H$ such as $M$'' leads to \textit{person}, \textit{election}, \textit{time}, \textit{thing}, \textit{leader} as the top five types. But using the pattern ``$M$ and any other $H$,'' we get \textit{candidate}, \textit{politician}, \textit{man}, \textit{person}, \textit{governor}.
On the other hand, for mention ``the Al Merreikh Stadium'' in ``It was only Chaouchi's third cap during that unforgettable night in the Al Merreikh Stadium,'' the results of using ``$H$ such as $M$'' (the top five types are \textit{venue}, \textit{place}, \textit{facility}, \textit{location}, \textit{area}) is better than using ``$M$ and any other $H$'' (the top five types are \textit{venue}, \textit{stadium}, \textit{game}, \textit{match}, \textit{time}).

To address the above problem, we do not use a same pattern for all the mentions. Instead, for each mention, we try to select the best pattern to apply from a list of patterns. This is achieved by using a baseline ultra-fine entity typing model, \textit{BERT-Ultra-Pre}, which is trained beforehand without using labels generated with our BERT MLM based approach. Details of \textit{BERT-Ultra-Pre} can be found in Section \ref{sec:ablation}. Denote the pattern list as $L$. With each pattern in $L$, we can apply it on the given mention to derive a set of labels from the BERT MLM. Then, we find the set of labels that have the most overlap with the labels predicted by \textit{BERT-Ultra-Pre}. Finally, the given mention is annotated with this set of labels.

It is not necessary to use all the patterns in \cite{seitner2016large}. 
To construct $L$, the list of patterns used for annotation, we perform the following procedure.

\vspace{0.1in}
\noindent\hangindent=1.2cm Step 1: Initialize $L$ to contain the best performing pattern (i.e., ``$M$ and any other $H$'') only. 

\noindent\hangindent=1.2cm Step 2: From all the patterns not in $L$, find the one that may bring the greatest improvement in F1 score if it is added to $L$.

\noindent\hangindent=1.2cm Step 3: Add the pattern found in Step 2 to the $L$ if the improvement brought by it is larger than a threshold.

\noindent\hangindent=1.2cm Step 4: Repeat steps 2-3 until no patterns can be added.

\paragraph{Discussion on Type Coverage} Since we only use one [MASK] token to generate labels, the model cannot produce multi-word types (e.g., \textit{football\_player}) or single word types that are not present in the BERT MLM vocabulary. The BERT MLM vocabulary covers about 92\% of the labels in the human annotated dataset constructed by \citet{choi2018ultra}.
Type coverage is a known issue with weak supervision, and is tolerable if the generated labels can be used to achieve our final goal: improving the performance of the ultra-fine entity typing model.

\subsection{Training Data}

Our approach generates type labels for all three types of mentions: named entity mentions, pronoun mentions, and nominal mentions.
For named entity mentions and nominal mentions, existing automatic annotation approaches can already provide some labels for them by using the entity types in knowledge bases or using the head words as types \cite{ling2012fine,choi2018ultra}. Thus, we combine these labels with the labels generated by us.
For pronoun mentions, no other labels are used.

Besides the automatically annotated samples, we can also use a small amount of human annotated samples provided by the dataset for model training.

\subsection{Model Training}
\label{sec:model-train}

Our ultra-fine entity typing model follows the BERT-based model in \cite{onoe2019learning}. Given a sentence that contains an entity mention, we form the sequence 
``[CLS] sentence [SEP] mention string [SEP]'' as the input to BERT.  
Then, denoting the final hidden vector of the ``[CLS]'' token as $\bm{u}$, we add a linear classification layer on top of $\bm{u}$ to model the probability of each type:
\begin{equation}
    \bm{p}=\sigma(\bm{W}\bm{u}),
\end{equation}
where $\sigma$ is the sigmoid function, $\bm{W}$ is a trainable weight matrix. $\bm{p} \in \mathbb{R}^d$, where $d$ is the number of types used by the dataset. We assign a type $t$ to the mention if $p_t$, its corresponding element in $\bm{p}$, is larger than 0.5. If no such types are found, we assign the one with the largest predicted probability to the mention.

To make use of the automatically labeled samples, some existing approaches mix them with high quality human annotated samples while training models \cite{choi2018ultra,onoe2019learning}. However, we find that better performance can be obtained by pretraining the model on automatically labeled samples, then fine-tuning it on human annotated samples.

Following \cite{choi2018ultra}, we partition the whole type vocabulary used by the dataset into three non-overlapping sets: general, fine, and ultra-fine types, denoted with $T_g,T_f$ and $T_u$, respectively. Then, we use the following objective for training:
\begin{equation}
\begin{split}
    \mathcal{J}(x)&=\mathcal{L}(x,T_g)\mathbbm{1}(\mathfrak{L}, T_g)+\mathcal{L}(x,T_f)\mathbbm{1}(\mathfrak{L}, T_f) \\
    &+\mathcal{L}(x,T_u)\mathbbm{1}(\mathfrak{L}, T_u),
\end{split}
\end{equation}
where $x$ is a training sample; $\mathfrak{L}$ denotes the set of type labels assigned to $x$ through either human or automatic annotation. The function $\mathbbm{1}(\mathfrak{L}, T)$ equals 1 when a type in $\mathfrak{L}$ is in set $T$ and 0 otherwise. This loss can avoid penalizing some false negative labels.

Unlike existing studies, we define the function $L$ differently for human annotated samples and automatically labeled samples.
While pretraining with automatically labeled samples, the labels obtained through entity linking and head word supervision are usually of higher precision than those obtained through BERT MLM. Thus, we propose to assign different weights in the training objective to the labels generated with different methods:
\begin{equation}
\label{eq:weight-loss}
\begin{split}
    \mathcal{L}(x,T)&=-\sum_{t\in T} \alpha(t) [y_t\cdot\log(p_t) \\
    &+(1-y_t) \cdot \log(1-p_t)],
\end{split}
\end{equation}
where $y_t$ equals to 1 if $t$ is annotated as a type for $x$ and 0 otherwise; $p_t$ is the probability of whether $t$ should be assigned to $x$ predicted by the model.
The value of $\alpha(t)$ indicates how confident we are about the label $t$ for $x$. Specifically, it equals to a predefined constant value larger than 1 when $t$ is a positive type for $x$ obtained through entity linking or head word supervision, otherwise, it equals to 1.

While fine-tuning with human annotated samples, we directly use the binary cross entropy loss:
\begin{equation}
    \mathcal{L}(x,T)=-\sum_{t\in T} [y_t\cdot\log(p_t)+(1-y_t) \cdot \log(1-p_t)].
\end{equation}


\subsection{Self-Training}
\label{sec:self-train}



Denote the ultra-fine entity typing model obtained after pretraining on the automatically labeled data as $h$, and the model obtained after fine-tuning $h$ with human annotated data as $m$. A weakness of $m$ is that at the fine-tuning stage, it is trained with only a small number of samples. Thus, we employ self-training to remedy this problem.

By using $m$ as a teacher model, our self-training step fine-tunes the model $h$ again with a mixture of the samples from the automatically labeled data and the human annotated data. This time, for the automatically annotated samples, we use pseudo labels generated based on the predictions of $m$ instead of their original weak labels. The newly fine-tuned model should perform better than $m$, and is used for evaluation.

Denote the set of human annotated samples as $H$, the set of automatically labeled samples as $A$. The training objective at this step is 
\begin{equation}
\label{eq:st-loss}
    \mathcal{J}_{ST}=\frac{1}{|H|}\sum_{x\in H}\mathcal{J}(x)+\lambda \frac{1}{|A|}\sum_{x\in A}\mathcal{L}_{ST}(x),
\end{equation}
where $\lambda$ is a hyperparameter that controls the strength of the supervision from the automatically labeled data.

While computing loss for the samples in $A$, we only use the types that are very likely to be positive or negative. For a sample $x$, let $p_t$ be the probability of it belonging to type $t$ predicted by the model $m$. We consider a type $t$ very likely to be positive if $p_t$ is larger than a threshold $P$, or if $t$ is a weak label of $x$ and $p_t$ is larger than a smaller threshold $P_w$. Denote the set of such types as $\hat{Y}^+(x)$. We consider a type $t$ very likely to be negative if $p_t$ is smaller than $1-P$. Denote the set of such types as $\hat{Y}^-(x)$. Then we have:
\begin{equation}
\begin{split}
    \mathcal{L}_{ST}(x)&=-\sum_{t\in \hat{Y}^+(x)} \log(p_t) \\
    &-\sum_{t\in \hat{Y}^-(x)} \log(1-p_t).
\end{split}
\end{equation}
Thus, we compute the binary cross entropy loss with only the types in $\hat{Y}^+(x)$ and $\hat{Y}^-(x)$.

\section{Application to Traditional Fine-grained Entity Typing}
\label{sec:fine-grained-app}

Our approach to generating weak entity type labels with BERT MLM can also be applied to the traditional fine-grained entity typing task. Different from ultra-fine entity typing, traditional fine-grained entity typing uses a manually designed entity type ontology to annotate mentions. The types in the ontology are organized in an hierarchical structure. For example, the ontology used by the Ontonotes dataset contains 89 types including /organization, /organization/company, /person, /person/politician, etc. On this dataset, our automatic annotation approach can mainly be helpful to generate better labels for nominal mentions. 

We still use the same method described in Section \ref{sec:bert-mlm-labels} to create input for BERT MLM based on the given mention. But with traditional fine-grained entity typing, most mentions are assigned only one type path (e.g., a company mention will only be assigned labels \{/organization, /organization/company\}, which includes all the types along the path of /organization/company). Thus, while generating labels, we only use the most probable word predicted by the BERT MLM, which is mapped to the types used by the dataset if possible. For example, the word ``company'' and its plural form are both mapped to /organization/company. Such a mapping from free-form entity type words to the types used by the dataset can be created manually, which does not require much effort. We mainly construct the mapping with two ways: 1) Check each type used by the dataset, and think of a few words that should belong to it, if possible. For example, for the type /person/artist/author, corresponding words can be ``author,'' ``writer,'' etc. 2) Run the BERT MLM on a large number of inputs constructed with unannotated mentions, then try to map the words that are most frequently predicted as the most probable word to the entity type ontology.

Since only the most probable word predicted by the BERT MLM is used to produce labels, we also only use one hypernym relation pattern: ``$M$ and any other $H$.''

For traditional fine-grained entity typing, we use our approach to generate labels for mentions that are not previously annotated with other automatic annotation approaches. While training, all the automatically labeled mentions are used together. The typing model is the same as the model described in \ref{sec:model-train}. The binary cross entropy loss is directly employed as the training objective.

\section{Experiments}
\label{sec:experiments}

We conduct experiments on our primary task: ultra-fine entity typing. In addition, we evaluate the performance of our approach when applied to traditional fine-grained entity typing.

\subsection{Evaluation on Ultrafine}

For ultra-fine entity typing, we use the dataset created by \citet{choi2018ultra}. It uses a type set that contains 10,331 types. These types are partitioned into three categories: 9 general types, 121 fine-grained types, and 10,201 ultra-fine types. There are 5,994 human annotated samples. They are split into train/dev/test with ratio 1:1:1. It also provides 5.2M samples weakly labeled through entity linking and 20M samples weakly labeled through head word supervision.

We compare with the following approaches:

\begin{itemize}
\item \textbf{UFET} \cite{choi2018ultra}. This approach  obtains the feature vector for classification by using a bi-LSTM, a character level CNN, and pretrained word embeddings.

\item \textbf{LabelGCN} \cite{xiong2019imposing}. LabelGCN uses a graph propagation layer to capture label correlations.

\item \textbf{LDET} \cite{onoe2019learning}. LDET learns a model that performs relabeling and sample filtering to the automatically labeled samples. Their typing model, which employs ELMo embeddings and a bi-LSTM, is train with the denoised labels.

\item \textbf{Box} \cite{onoe2021modeling}. \textit{Box} represents entity types with box embeddings to capture latent type hierarchies. Their model is BERT-based.

\end{itemize}



We use the BERT-Base-Cased version of BERT for both weak label generation and the typing model in Section \ref{sec:model-train}. The hyperparameters are tuned through grid search using F1 on the dev set as criterion. The value of $\alpha(t)$ in Equation (\ref{eq:weight-loss}) is set to 5.0 for positive types obtained through entity linking or head word supervision. $\lambda$ in Equation (\ref{eq:st-loss}) is set to 0.01. $P$ and $P_w$ in Section \ref{sec:self-train} are set to 0.9 and 0.7, respectively. Our approach to generate labels through BERT MLM is applied to each weak sample provided in the original dataset. In addition, we also use our approach to annotate about 3.7M pronoun mentions, which are extracted through string matching from the English Gigaword corpus \cite{parker2011english}. We generate 10 types for each sample\footnote{The performance of the trained model is relatively insensitive with respect to the number of labels generated with MLM. The difference between the F1 scores of the models trained using 10 and 15 generated types is less than 0.005.}. With the procedure described in Sectiton \ref{sec:bert-mlm-labels}, three hypernym extraction patterns are used while generating labels with BERT MLM: ``$M$ and any other $H$,'' ``$H$ such as $M$,'' ``$M$ and some other $H$.'' Specifically, adding ``$H$ such as $M$'' and ``$M$ and some other $H$'' improves the F1 score from 0.253 to 0.274, and from 0.274 to 0.279, respectively. Adding any more patterns cannot improve the F1 score for more than 0.007.

\begin{table}
\centering
\begin{tabular}{lccc}
\hline \textbf{Method} & \textbf{P} & \textbf{R} & \textbf{F1} \\ \hline
UFET & 47.1 & 24.2 & 32.0 \\
LabelGCN & 50.3 & 29.2 & 36.9 \\
LDET & 51.5 & 33.0 & 40.2 \\
Box & 52.8 & 38.8 & 44.8 \\ \hline
Ours & \textbf{53.6} & \textbf{45.3} & \textbf{49.1} \\ \hline
\end{tabular}
\caption{\label{tab:performance-main} Macro-averaged Precision, Recall, and F1 of different approaches on the \textit{test set}.}
\end{table}

Following existing work \cite{onoe2021modeling,onoe2019learning}, we evaluate the macro-averaged precision, recall, and F1 of different approaches on the manually annotated test set. The results are in Table \ref{tab:performance-main}. Our approach achieves the best F1 score. It obtains more than 4\% F1 score improvement over the existing best reported performance by Box in \cite{onoe2021modeling}. This demonstrates the effectiveness of our approach.



\subsection{Ablation Study}
\label{sec:ablation}

\begin{table}
\centering
\begin{tabular}{lccc}
\hline \textbf{Method} & \textbf{P} & \textbf{R} & \textbf{F1} \\ \hline
BERT-Ultra-Direct & 51.0 & 33.8 & 40.7 \\ 
BERT-Ultra-Pre & 50.8 & 39.7 & 44.6 \\
Ours (Single Pattern) & 52.4 & 44.9 & 48.3 \\
Ours (Unweighted Loss) & 51.5 & 45.8 & 48.5 \\
Ours (No Self-train) & 53.5 & 42.8 & 47.5 \\
Ours & \textbf{53.6} & \textbf{45.3} & \textbf{49.1} \\ \hline
\end{tabular}
\caption{\label{tab:performance-abl} Performance of different variants of our approach on the \textit{test set}. BERT-Ultra-Direct and BERT-Ultra-Pre are two baseline approaches that do not use labels generated with our BERT MLM based method in training.
}
\end{table}

For ablation study, we verify the effectiveness of the different techniques used in our full entity typing approach by evaluating the performance of the following variants: \textbf{Ours (Single Pattern)} only uses one pattern: $M$ and any other $H$; \textbf{Ours (Unweighted Loss)} removes the $\alpha(t)$ term in Equation (\ref{eq:weight-loss}); \textbf{Ours (No Self-train)} does not perform the self-training step. We also evaluate two baseline approaches: \textbf{BERT-Ultra-Direct} uses the same BERT based model described in Section \ref{sec:model-train}, but is trained with only the human annotated training samples; \textbf{BERT-Ultra-Pre} also uses the same BERT based model, but is first pretrained with the existing automatically generated training samples in the dataset provided by \citet{choi2018ultra}, then fine-tuned on the human annotated training data.


First, the benefit of using the labels generated through BERT MLM can be verified by comparing \textit{Ours (No Self-train)} and BERT-Ultra-Pre. Because the techniques employed in \textit{Ours (No Self-train)}, including the use of multiple hypernym extraction patterns and the weighted loss, are both for better utilization of our automatic entity type label generation method.

The effectiveness of the use of multiple hypernym extraction patterns, the weighted loss, and the self-training step can be verified by comparing \textit{Ours} with \textit{Ours (Single Pattern)}, \textit{Ours (Unweighted Loss)} and \textit{Ours (No Self-train)}, respectively. Among them, self-training is most beneficial.





\subsection{Evaluation on Different Kinds of Mentions}

\begin{table*}
\centering
\begin{tabular}{lccccccccc}
\hline & \multicolumn{3}{c}{Named Entity} & \multicolumn{3}{c}{Pronoun} & \multicolumn{3}{c}{Nominal} \\
\hline \textbf{Method} & \textbf{P} & \textbf{R} & \textbf{F1} & \textbf{P} & \textbf{R} & \textbf{F1} & \textbf{P} & \textbf{R} & \textbf{F1} \\ \hline
BERT-Ultra & 58.1 & 45.1 & 50.8 & 52.9 & 42.9 & 47.4 & 47.4 & 26.9 & 34.3 \\
BERT-Ultra-Pre & 54.7 & 50.5 & 52.5 & 51.3 & 46.1 & 48.6 & 45.2 & 33.7 & 38.6 \\
Ours & 58.3 & 54.4 & 56.3 & 57.2 & 50.0 & 53.4 & 49.5 & 38.9 & 43.5 \\ \hline
\end{tabular}
\caption{\label{tab:performance-mkind} Performance on named entity mentions, pronoun mentions, and nominal mentions, respectively.}
\end{table*}

\begin{table}[t]
\centering
{\small
\begin{tabular}{lp{0.28\textwidth}}
\hline 
\textbf{Sentence} &
\begin{tcolorbox} Captured in 1795, \underline{\textbf{he}} was confined at Dunkirk, escaped, set sail for India, was wrecked on the French coast, and condemned to death by the decree of the French Directory.\end{tcolorbox} \\
\textbf{Human} & prisoner, person \\
\textbf{BERT-Ultra-Pre} & person, soldier, man, criminal \\
\textbf{BERT MLM} & man, prisoner, person, soldier, officer \\
\textbf{Ours} & person, soldier, man, prisoner \\
\hline
\textbf{Sentence} & \begin{tcolorbox} Also in the morning, \underline{\textbf{a roadside bomb}} struck a police patrol on a main road in Baghdad's northern neighborhood of Waziriya, damaging a police vehicle ...\end{tcolorbox} \\
\textbf{Human} & bomb, weapon, object, explosive \\
\textbf{BERT-Ultra-Pre} & object, event, attack, bomb \\
\textbf{BERT MLM} & weapon, threat, evidence, device, debris \\
\textbf{Ours} & object, weapon, bomb \\
\hline
\textbf{Sentence} & \begin{tcolorbox} In October 1917, \underline{\textbf{Sutton}} was promoted (temporarily) to the rank of major and appointed Officer Commanding No.7 Squadron, a position he held for the remained of the War.\end{tcolorbox} \\
\textbf{Human} & soldier, officer, male, person \\
\textbf{BERT-Ultra-Pre} & person, politician, male \\
\textbf{BERT MLM} & officer, pilot, man, unit, aircraft \\
\textbf{Ours} & person, soldier, male, officer \\
\hline
\end{tabular}
}
\caption{\label{tab:cases} Ultra-fine entity typing examples with the corresponding human annotated labels and predictions of three different systems. Entity mentions are in bold and underlined. For BERT MLM, we list the top five labels.}
\end{table}

It is also interesting to see how our approach performs on different kinds of mentions. 
Table \ref{tab:performance-mkind} lists the performance of our full approach and two baseline systems on the three kinds of mentions in the dataset: named entity mention, pronoun mentions, and nominal mentions.

Our approach performs much better than BERT-Ultra-Pre on all three kinds of mentions. The improvements in F1 on pronoun and nominal mentions are relatively more substantial.


\subsection{Case Study}

Table \ref{tab:cases} presents several ultra-fine entity typing examples, along with the human annotated labels, and the labels predicted by BERT-Ultra-Pre, BERT MLM, and our full approach.


In the first example, the label \textit{prisoner} is a type that depends on the context, and is usually not assigned to humans in knowledge bases. We think that since we can assign such labels to the training samples with our BERT MLM based approach, our model is better at predicting them than the baseline model.

The second and third examples demonstrate that our model may not only improve the recall by predicting more correct types, but also reduce incorrect predictions that do not fit the mention or the context well.

\subsection{Evaluation on Ontonotes}

The Ontonotes dataset uses an ontology that contains 89 types to label entity mentions. We use the version provided by \citet{choi2018ultra}. It includes 11,165 manually annotated mentions, which are split into a test set that contains 8,963 mentions, and a dev set that contain 2,202 mentions. It also provides about 3.4M automatically labeled mentions.

Since existing annotations for named entity mentions may be more accurate than the annotations obtained through our approach, we only apply our method to label nominal mentions. Applying the approach in Section \ref{sec:fine-grained-app}, we create 1M new automatically labeled mentions with the head word supervision samples (such samples contain mostly nominal mentions) in the ultra-fine dataset. They are used together with the originally provided 3.4M mentions to train the typing model.

On this dataset, we compare with the following approaches: \textbf{UFET} \cite{choi2018ultra}, \textbf{LDET} \cite{onoe2019learning}, \textbf{DSAM} \cite{hu2020diversified}, \textbf{LTRFET} \cite{lin2019attentive}, \textbf{BERT-Direct}. Where BERT-Direct uses the same BERT based model as our approach, but trains with only the weak samples provided in the dataset. 
LTRFET adopts a hybrid classification method to exploit type inter-dependency. DSAM is a diversified semantic attention model with both mention-level attention and context-level attention.

For our approach and BERT-Direct, we still use the pretrained BERT-Base-Cased model for initialization. Although a very large number of weakly labeled mentions are provided, not all of them are needed for training the models. In our experiments, for both our approach and BERT-Direct, the performance does not increase after training on about 0.3M mentions.

\begin{table}[t]
\centering
\begin{tabular}{lccc}
\hline \textbf{Method} & \textbf{Acc} & \textbf{Macro F1} & \textbf{Micro F1} \\ \hline
UFET & 59.5 & 76.8 & 71.8 \\ 
LTRFET & 63.8 & 82.9 & 77.3 \\ 
LDET & 64.9 & 84.5 & 79.2 \\ 
DSAM & 66.06 & 83.07 & 78.19 \\ \hline
BERT-Direct & 63.25 & 80.84 & 75.90 \\
Ours & \textbf{67.44} & \textbf{85.44} & \textbf{80.35} \\ \hline
\end{tabular}
\caption{\label{tab:performance-ontonotes} Performance of different approaches on Ontonotes. We report strict accuracy, macro-averaged F1, and micro-averaged F1.}
\end{table}

We report strict accuracy, macro-averaged F1, and micro-averaged F1 \cite{ling2012fine}. The results are in Table \ref{tab:performance-ontonotes}. As we can see, our approach also achieves the best performance on this dataset.
Comparing it with BERT-Direct demonstrates the benefit of the samples automatically labeled with BERT MLM.

However, less improvement is achieved on OntoNotes than on the ultra-fine entity typing dataset. We think there are two main reasons. First, OntoNotes uses a much smaller entity type set (89 types) than the ultra-fine entity typing dataset (10,331 types). As a result, some finer grained types that can be produced by our approach become less beneficial. Second, generating type labels that are highly dependent on the context (e.g., types like \textit{criminal}, \textit{speaker}) is an advantage of our approach, and the ultra-fine entity typing dataset contains more such type labels. 

\section{Conclusion}

In this work, we propose a new approach to automatically generate ultra-fine entity typing labels. Given a sentence that contains a mention, we insert a hypernym extraction pattern with a ``[MASK]'' token in it, so that a pretrained BERT MLM may predict hypernyms of the mention for ``[MASK].''  Multiple patterns are used to produce better labels for each mention. We also propose to use a weighted loss and perform a self-training step to learn better entity typing models. Experimental results show that our approach greatly outperforms state-of-the-art systems. Additionally, we also apply our approach to traditional fine-grained entity typing, and verify its effectiveness with experiments.

\section*{Acknowledgments}

This paper was supported by the NSFC Grant (No. U20B2053) from China, the Early Career Scheme (ECS, No. 26206717), the General Research Fund (GRF, No. 16211520), and the Research Impact Fund (RIF, No. R6020-19 and No. R6021-20) from the Research Grants Council (RGC) of Hong Kong, with special thanks to the WeChat-HKUST WHAT Lab on Artificial Intelligence Technology.


\bibliographystyle{acl_natbib}
\bibliography{acl2021}


\end{document}